\documentclass{article}

\usepackage{arxiv}

\usepackage[utf8]{inputenc} % allow utf-8 input
\usepackage[T1]{fontenc}    % use 8-bit T1 fonts
\usepackage{hyperref}       % hyperlinks
\usepackage{url}            % simple URL typesetting
\usepackage{booktabs}       % professional-quality tables
\usepackage{amsfonts}       % blackboard math symbols
\usepackage{nicefrac}       % compact symbols for 1/2, etc.
\usepackage{microtype}      % microtypography
\usepackage{lipsum}
\usepackage{graphicx}
\graphicspath{ {./images/} }

\title{San-BERT: Extractive Summarization for Sanskrit Documents using BERT and its variants}

\author{
 Kartik Bhatnagar \\
  Department of Mathematics and Computer Science\\
  Sri Sathya Sai Institute of Higher Learning \\
  Prashanti Nilayam, Andra Pradesh, 
  IN 515134\\
  \texttt{kartikbhatnagar15@gmail.com} \\
   \And
 Sampath Lonka \\
  Department of Mathematics and Computer Science\\
  Sri Sathya Sai Institute of Higher Learning \\
  Prashanti Nilayam, Andra Pradesh, 
  IN 515134\\
  \texttt{sampathlonka@sssihl.edu.in} \\
  \And
 Jammi Kunal \\
  Department of Mathematics and Computer Science\\
  Sri Sathya Sai Institute of Higher Learning \\
  Prashanti Nilayam, Andra Pradesh, 
  IN 515134\\
  \texttt{jammikunal000@gmail.com} \\
 \And
   Mahabala Rao M G \\
  Department of Language and Literature\\
  Sri Sathya Sai Institute of Higher Learning \\
  Prashanti Nilayam, Andra Pradesh, 
  IN 515134\\
  \texttt{mahabalaraomg@sssihl.edu.in} \\
}

\begin{document}
\maketitle
\begin{abstract}
In this work, we develop language models for the Sanskrit language, namely Bidirectional Encoder Representations from Transformers (BERT) and its variants: A Lite BERT (ALBERT), and Robustly Optimized BERT (RoBERTa) using Devanagari Sanskrit text corpus. Then we extracted the features for the given text from these models. We applied the dimensional reduction and clustering techniques on the features to generate an extractive summary for a given Sanskrit document. Along with the extractive text summarization techniques, we have also created and released a Sanskrit Devanagari text corpus publicly.
\end{abstract}

% keywords can be removed
\keywords{Sanskrit Documents \and Extractive Summarization \and Word Embedding\and Clustering \and ROUGE Score \and BERT Score}

\section{Introduction}
%Sanskrit is the language of scholars including Bhramagupta, Aryabhatta, Panini, Bhaskaracharya, etc. 
Sanskrit is an ancient and incredibly important language that has had an incredible impact on Indian literature. After all, it is the source of a vast body of scientific and literary works covering astronomy, mathematics, medicine, philosophy, technology, and so much more. Language was used as the main form of communication for scholarly discourses across India up until quite recently. As a result, Sanskrit literature has kept up a steady production of work throughout its two-thousand-year history \cite{amba2012}. Much of this literature has not yet been explored by modern humans.  
There is a wealth of knowledge stored in Sanskrit texts and ancient manuscripts that the public is unaware of. When information is abundant everywhere, people rarely have time to read a complete document in today's society. Having a summary of the content is preferable to avoid the tedious process of reading the entire document. An excellent summary can obtain crucial information and key points from the entire document, saving the reader time. Today, various summarization strategies are accessible, ranging from rule-based, statistical-based, graph-based, and current state-of-the-art deep learning-based techniques. Many Natural-Language-based jobs are improving due to recent advancements in the field of Deep Learning, and summarization is one of them.
%Most of these are data-driven approaches and these are not well-studied for low-resourced languages like Sanskrit. 
There are primarily two types of summarization techniques: abstractive and extractive.  In abstractive text summarization, we summarize the crucial information by paraphrasing all of the relevant statements in the document. In extractive text summarization, on the other hand, we extract the important sentences from a document and make it a summary. The extractive summary condenses all significant sentences that address the documents underlying theme and presents them in the form of a summary. Although people normally produce summaries in abstract ways, extractive summarization algorithms have received greater attention in recent studies \cite{Allahyari2017}. Extractive summarizing systems perform better than abstractive summarizing systems in many cases \cite{Erkan2004} based on available evaluation metrics.
We have the following findings in this work: 
\begin{enumerate}
    \item[(a)] Developed a digitalized Devanagari Sanskrit corpus and made the data available to the public\footnote{https://www.kaggle.com/datasets/kartikbhatnagar18/sanskrit-text-corpus}.
    \item[(b)] We have trained a set of language models for the Sanskrit language, including BERT, ALBERT and RoBERTa. There are different word-embedding techniques discussed for Sanskrit in \cite{Sandhan2021} using transliteration of Sanskrit text corpus. As far as we know, ours is the first attempt to develop BERT and its variants using the Devanagari script for Sanskrit.
    \item[(c)] We developed an extractive summary through the extraction of features from language models. Following that, we evaluated the generated summary utilizing ROUGE scores \cite{Chin-2004} and BERT scores \cite{Tiyani2019}.
\end{enumerate}
Most of these results are part of the first author's Master Dissertation \cite{karthik2022}.
\section{Previous Works}

Luhn H.P \cite{Luhn1958} initially presented a concept behind text summarization in 1958, based on word frequency. A further refinement of his ideas was proposed in \cite{Ed1969}, which suggested that sentence position and word order should be considered when assessing the importance of sentences to form an effective summary.  Since the 1970s, researchers have been developing rule-based approaches for summarization, but the problem they faced was that the rules were not generic; for different domain documents, they had to make different rules for summarization algorithms. In the 1990s, several graph-based techniques emerged, but they struggled to account for linguistic features. Due to the furtherance in computer linguistics, modern machine learning, and deep learning, text summarization has made immense strides in recent decades.

There are different techniques developed for text summarization both abstractive and extractive using machine learning models. As these models are data-driven, supervised \cite{Mandal2021},\cite{Nallapati2016},\cite{Yang2019}, unsupervised  \cite{Ren2017}, \cite{Erkan2004}, \cite{Moratanch2017}, \cite{Dutta2019} and Reinforcement Learning techniques \cite{narayan2018}, \cite{Kaichun2018} developed to generate the extractive summary. The supervised approach to summarizing requires a labeled data set. For Sanskrit, it can often be hard to find labeled data suitable for training a model. In this situation, an unsupervised approach is extremely advantageous.

To our knowledge, Barve et al. \cite{Barve2015} were the first researchers to generate extractive summaries from Sanskrit text. They applied various unsupervised query-based summarization approaches, namely, Vector Space Models (VSM), Graph-based methods using PageRank, and term-frequency and inverse-sentence frequency (tf-isf). Using these approaches extracted the sentences which are more relevant to the given query and ranked all the extracted sentences in descending order. The top-ranked sentences together form an extractive summary. Then, the generated summaries are evaluated by human experts in Sanskrit, and the results are shown in \cite{Barve2015}. 

Note that the results shown in both works \cite{Barve2015}, \cite{Shalini2021} are limited to a single page or small paragraphs.

\section{Language Models for Sanskrit}
Bidirectional Encoder Representations from Transformers \cite{Vaswani2017} (BERT) \cite{Devlin2019}, models have achieved state-of-the-art performance for many natural language processing (NLP) tasks. To use BERT models for various NLP tasks in Sanskrit, we will need to train the appropriate BERT family model(s). As such, a data-driven approach is taken and it is critical to have large corpora to support these models during training and testing. In this section, we discuss the training process required to effectively implement various BERT models.

\subsection{Data}
We scraped Sanskrit data in Devanagari script from different resources \footnote{https://sa.wikipedia.org/wiki/} \footnote{https://oscar-corpus.com/post/oscar-v22-01/} \footnote{https://www.sanskritworld.in/} \footnote{SanskritDocuments.org} \footnote{http://sanskrit.jnu.ac.in/currentSanskritProse/} \footnote{https://sanskrit.uohyd.ac.in/Corpus/}
we used Python packages Beautiful Soup 4.9.0 and requests 2.27.1 to extract the data from HTML pages. In addition, we remove extra spaces, extra punctuation marks, website links, and unwanted characters other than the Devanagari script from the scraped data using functions from the Regex Python library. After cleaning the document, the entire text present is split on the basis of the delimiter purnavirama ('$|$'), resulting in creating the list of sentences. After cleaning, we have 800 MB of Sanskrit Text Corpus (STC) for our training purpose and it is available for public \footnote{https://www.kaggle.com/datasets/kartikbhatnagar18/sanskrit-text-corpus}.  We have a total of 30M tokens in the STC.

\subsection{Training Language Models}
The architecture and configurations of the models  BERT, ALBERT, and RoBERTa are taken from Hugging Face\footnote{https://huggingface.com/models}.

\subsubsection{San-BERT}
\textit{Bidirectional Encoder-Representation from Transformer} (BERT) \cite{Devlin2019} is a transformer-based
natural language processing model that uses surrounding text to help computers understand what words imply in context. Unlike other standard language models, which can only
read sequential data in one direction, the BERT model can read data in both directions, from
left to right and vice versa, resulting in improved learning. There are majorly two tasks on
which BERT Model is prepared, one is Masked-Language Model (MLM), and the other is
Next-Sentence Prediction (NSP). In the MLM task, throughout the text data, few words are
masked, and BERT tries to predict those words, and in the NSP task, BERT tries to find out whether the preceding line is logical or correct. These two tasks help the model learn the
inner parameters, which are used to perform downstream tasks. For fine-tuning BERT for our corpus, we use only the MLM task. For the training BERT, 
since the NLP community avoids BERT models to train on NSP task, as it is an easy task for the
model to in comparison to the MLM task, and it does help the model to learn better, so
following the same line we trained our BERT model in MLM task.
The first task during the training of the BERT model is preparing the vocabulary; generally,
BERT models have around 30k tokens in their vocabulary, which are sufficient enough to
represent any text data for training. To create our Vocabulary for Sanskrit text we used BERT
Word Piece Tokenizer, the tokenizers have the capability to deal with the problem of Out Of
Vocabulary Word (OOV) Problem. The tokenizer keeps the whole word in the vocabulary,
or in some cases a whole word is divided into different parts; in the latter case, the broken
word is prefixed with a double hash($\#\#$), symbolizing it as part of the whole word. So, if in
case a larger word comes to the model which is not directly represented in the vocabulary,
then in this case, to deal with the problem of out-of-vocabulary (OOV), the tokenizer will simply break the words into smaller pieces until it matches its sub-word kept prefixed with double hash($\#\#$).The use of creating the vocabulary is that we can represent a text (line or paragraph) to its equivalent token id, as the computer would not accept a word directly; rather it would be suitable to represent a word in its equivalent mathematical form. The input taken by a BERT model for training is text data in its equivalent token ids (input ids), attention masks, and special token masks. For our task, we have trained the BERT model on our data set with MLM task, on Lenovo Legion RTX 3060 6 GB Nvidia Graphic card, CPU processor AMB 5800 H, SSD 256 GB for 18 hours.

\subsubsection{San-ALBERT}
BERT models have revolutionized the field of NLP with their outstanding results; the disadvantage of BERT models is their giant architecture size. For a BERT model, we need to train about 340M parameters, and it is a huge task to do it on a personal computer. In order to reduce the computational parameters and the training time, we will be using another variant of the BERT model, that is, a Lite BERT (ALBERT) \cite{Lan2019}. Using the novel techniques ALBERT model reduces around 90 percent of the parameters in comparison to the BERT model and it still can maintain its performance and to scale up to the level of BERT. The tokenizer used for training the vocabulary is Sentence-Piece Trainer. With the help of a Sentence-Piece trainer, around 32k tokens were produced. These tokens were constituents of words in their full form, subwords prefixed with a double hash ($\#\#$), and some special tokens such as [CLS], [SEP], [MASK]. After making the vocabulary, we imported the ALBERTTokenizer function to tokenize throughout the corpus, which will be given for training. The same Masked Language Model (MLM) training task was used to train the ALBERT model, around 15\% of the words were masked for the training. We trained our ALBERT model on the Google Colaboratory, Tesla K-80 GPU with driver version
460.32.03 and CUDA version 11.2 for about 9 hours. Around 80 percent of the text was in the training file and around 20 percent of the text data was in the validation file.

\subsubsection{San-RoBERTa}
Robustly optimized BERT (RoBERTa) \cite{Liu2019} is another variant of the BERT model; it was developed by Meta AI researchers; they thoroughly examined the BERT architecture, which was developed by Google, and found the various optimizations to be done in the architecture to perform high quality related to natural languages. In RoBERTa, there has been an improvement about the hyperparameter and the learning rate is also comparatively higher, which helps the model to learn faster. The Next Sentence Prediction (NSP) task is removed in the RoBERTa Model and the Masked Language Modeling (MLM) task is improved. With such advantages being offered by the RoBERTa Model, we also trained a RoBERTa model on our corpus. The tokenizer that we used to make the vocabulary is the Byte Level tokenizer namely \emph{Bpet RoBERTa} tokenizer. We made a vocabulary of around 30K tokens and send the whole corpus through the tokenizer. The RoBERTa tokenizer is different from other tokenizers, it uses byte level tokenizer which is also used in GPT-2 \cite{radford2019} architectures, and the schemas used in pertaining techniques are different, which helps the model to give better representation for training. The special tokens used by the tokenizer are: $<s>,\ <pad>,\ </s>,\ <unk>,\ <mask>.$

From the Hugging Face\footnote{https://huggingface.co/docs/transformers/model$\_$doc/roberta}, we imported RobertaTokenizer, RobertaConfig, and RobertaForMaskedLM for our training purpose. We used PyTorch to represent labels, mask tokens, input ids, and mask arrays in the form of tensors, which are then fed to the input layer of the RoBERTa Model. In the input layer we adjusted the max length of the tensor to 514, the hidden layer size is 768, and the number of hidden layers is set to 6. The learning rate we used for training is $1e-4.$ We trained this model on an Nvidia Tesla P100 GPU and trained the model for around 18 hours.

\subsection{Embedding Representation}
The purpose of training the BERT, ALBERT, and RoBERTa models was to get a better-contextualized representation embedding of any given Sanskrit word. Contextualized embedding generates a separate representation for each sense of meaning depending on the context words surrounding \cite{Sandhan2021}.  These embeddings will be the representation of the words present in the document, for which we have to obtain the extractive summary.

\begin{figure}[ht]
\centering
\includegraphics[width=\linewidth]{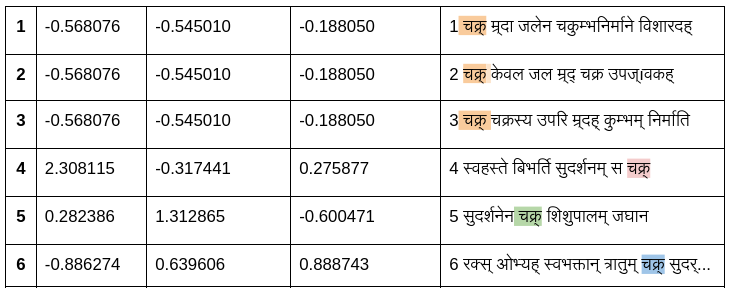}
\caption{Vector representation of the word $chakr$ in a 3-dimensional space using Contextualized word Embedding}
\label{fig4}
\end{figure}

To demonstrate the performance of our San-BERT model, we ran embedding for six sentences taken from the analysis of \cite{Sandhan2021}. After applying dimension reduction of output to three dimensions, the results are shown in fig.\ref{fig4}. In the first three examples, the Word $chakr$ means the $Potter$ and has the same word embedding, and in the following three sentences the word $chakr$ has a different meaning and so the embedding is different. 
For any given document we will be getting the embeddings in two stages, the first is Word Embedding and the other is Sentence Embedding.

\section{Extractive Text Summarization Techniques}
In this section, we explore several extractive summarization techniques to generate the summary from Sanskrit documents. We discussed two approaches to get the extractive summary for given Sanskrit document(s): statistical-based and neural-based. The details of each technique are discussed below:
\subsection{Statistical Based Technique}
The Term Frequency - Inverse Document Frequency (TF-IDF) is a statistical technique, which helps to find the significance of a given word in the entire document. TF-IDF calculates the score of a given word based on its repetitive occurrences in a document and also holds into account the word's occurrence in the other documents. 
TF-IDF  will help to capture the importance of words throughout the document by calculating the word occurrence in a particular sentence (TF) and also by calculating the number of sentences constituting that word (IDF). Together, TF and IDF, we calculate the TF-IDF score for each word $w$ as follows:
\[TF-IDF(w) = TF(w)*IDF(w)\]
We use these scores for each word in a given sentence $S,$ then we compute the sentence score \emph{Sentence TF-IDF Score} as follows:
\[Sentence\ TF-IDF\ Score = {\sum_{w\in S}TF-IDF(w)\over |S|}\]

\begin{centering}
\begin{figure}[ht]
\[\includegraphics[scale=0.65]{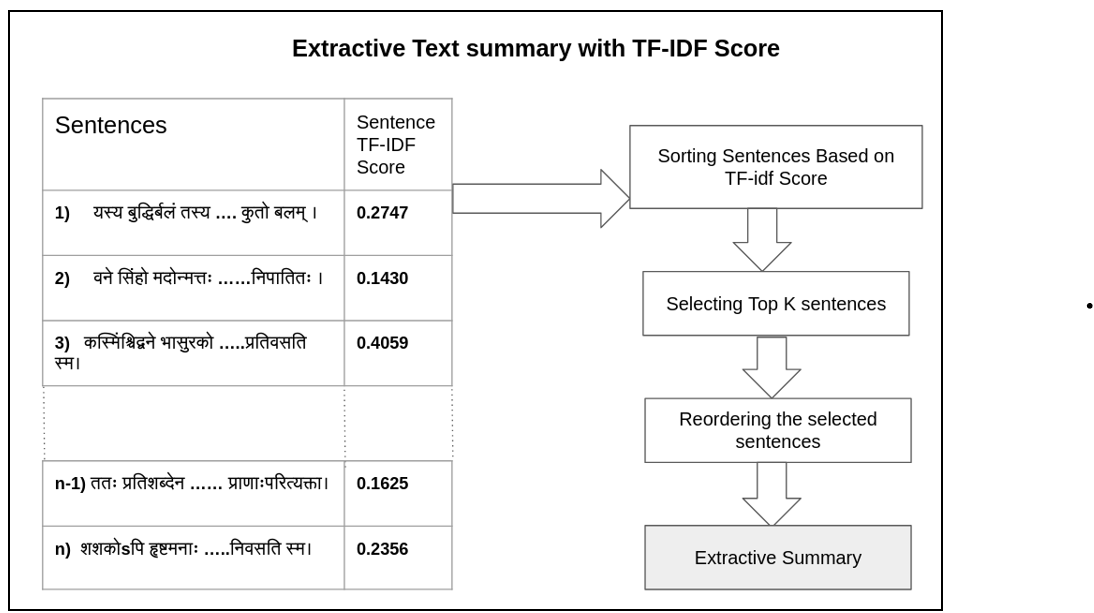}\]
\caption{Pipeline for Extractive text Summary using TF-IDF Scores}
\label{fig_tf}
\end{figure}
\end{centering}

We sort the sentences based on sentence TF-IDF Scores, then select the first $k$ sentences. We generate the extractive summary by reordering the top $k$ sentences. The procedure is shown in Figure \ref{fig_tf}.

\subsection{Neural Based Techniques}
In this subsection, we discuss how to generate an interactive summary using the BERT family of models.
\subsubsection{Overall Workflow}
For generating the extractive summary from the Sanskrit document(s), we use the preprocessed document(s). The list of sentences from the documents is passed through the model, where we will obtain the word embedding from the trained BERT models, and sentence embedding is obtained by averaging all the word embedding present in a sentence. Then with the help of the K-means clustering technique, we will make the clusters of sentence embedding vectors, then we will figure out the number of optimal clusters formed by the embeddings through the elbow method. After obtaining the number of clusters formed, the next task is to find out the central embedding in those clusters. In the next step, we will find the distance of each sentence embedding with the respective cluster's central embedding via a cosine distance measure. After having all the distances of sentences, we will sort the sentences according to the distances, the sentences having less distance from the central sentence will have a higher rank than the sentence far away from the center of the cluster. After ranking the sentences, the next step will be to select the top $k$ sentences to be displayed in summary. The next step after selecting the top - k-sentences is to reorder them back in chronological order. These selected important sentences will be the extractive summary of the whole document(s).

\begin{figure}[ht]
\centerline{\includegraphics[width=\linewidth]{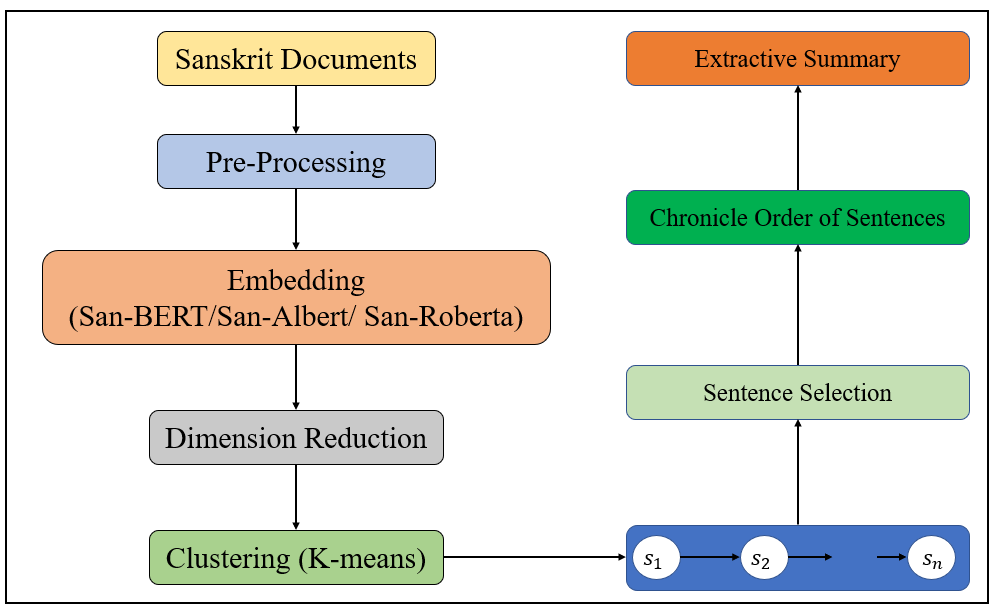}}
\caption{Pipeline for Extractive text Summary using variants of BERT Models}
\label{fig}
\end{figure}
\subsubsection{Token Embeddings}
In the BERT model, ALBERT model and RoBERTa model we have 12 encoders, 1 encoder and 6
encoders layers, respectively. The encoder layer is followed by the output layer, this layer
gives out the embedding of each token present in the input. The number of embeddings
coming out of the model is equal to the number of tokens produced by the tokenizer for a
given piece of input text.
\subsubsection{Sentence Embedding}
For a given document(s) our input to the BERT models will be in the form of sentences. These
sentences will be obtained from the preprocessing and data cleaning stage. Then for each
sentence passed to a BERT model we will be getting different embeddings for each token
present in the sentence. In order to get the Sentence embedding, we will average out all the
token embeddings obtained. 
\subsection{Dimension Reduction}
The Embeddings of a single sentence we are obtaining from the BERT models are of very high
dimension (768), and in a document(s) we will have numerous sentences. There are various disadvantages to using data in higher dimensions, the foremost of which are high data storage, and high data computation. The distances between the closest and farthest data
points can become equidistant in high dimensions, limiting the accuracy of some
distance-based analytic techniques.
To deal with the problems given by the higher-dimensional data, the best option is to reduce
the data dimension. Dimensionality reduction aids in the resolution of these
above-mentioned challenges while striving to preserve the majority of the important data.
The more features there are in the data it will be difficult to visualize and then operate on the
training set. Most of these traits are frequently linked and thus redundant. In this
case, dimensionality reduction techniques are useful. The dimension reduction we are going
to use is PCA ( Principal-Component Analysis), which discovers the relationships between the data.
PCA in the high-dimensional data finds the highest-variance directions and then projects
them onto a new subspace with fewer dimensions in comparison to the original data. Thus
PCA will be helpful to us to reduce the Sentence embedding dimensions and work
efficiently with the sentence embedding in the lower dimension space.
\subsection{Clustering}
A whole document(s) will not be a constituent of the same topic or theme, within a document
there will be different subtopics and different themes present in it. So to capture
capturing all the themes and subtopics within a document(s), in an Extractive Summary, we
will be using the Clustering technique.
 
\begin{figure}[ht]
\centering
\includegraphics[scale=1]{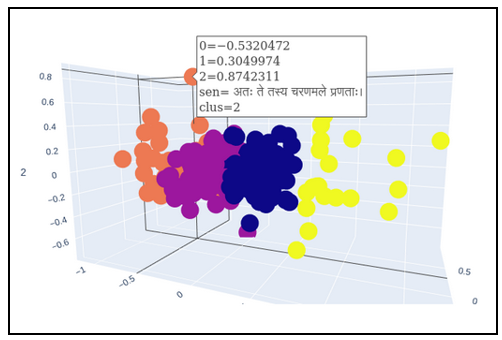}
\caption{Different Clusters Formed by the Sentences Embedded within a Document }
\label{fig3}
\end{figure}

After obtaining the sentence embeddings of all the sentences present throughout the document(s), we will find the optimal number of clusters formed by those embeddings with the help of the Elbow method. Based on our experiments on average we are getting
around the optimal 3 to 4 clusters for a document. After finding out the optimal number of
clusters, we will use the K-means clustering technique and assign sentences to its cluster.
Sentences present in the same cluster represent the same subtopic or theme of the
document. Thus considering this idea will give the extractive summary consisting of the main
points of each sub-topic and theme.
\subsection{Sentence Ranking}
In the next step after assigning each sentence to its cluster, we will be finding out the distance
of each embedding from the center of its cluster. We used various distance measures like
Manhattan distance, Euclidean Distance, Cosine similarity measure, and cosine distance for finding out the distances. It turned out that cosine similarity and cosine distance measures
were giving more accurate results. So, based on the cosine similarity and cosine distances we
will rank the sentences. The sentence embedding whose cosine distance from the center of
its cluster is minimal will have a higher rank and those sentence embeddings, with a cosine
distance is more will have a poor rank. Following this procedure, we will have all the sentences
sorted based on their ranks.        
\subsection{Extractive Summary}
After obtaining the sentences according to their ranks, the next task is to select the top $k$
sentences that are going to be part of the final extractive summary. We made our function
dynamic, users can select the number of lines they want in their Extractive Summary. It is
generally preferred to have 1/5th of the total lines as a summary, as it can cover all the
aspects of the document(s) in a condensed form. After selecting the top sentences, the next
process is to reorder them back in the chronological order of their occurrence.
Our neural-based models are capable of generating a summarization of any Sanskrit documents in Devanagari script and condensing essential, relevant information into one concise summary. 

\section{Results}
In this section, we evaluate our models and present the results. We evaluated our models in both cases: smaller documents (10 - 15 sentences) and longer documents (A couple of pages).

\subsection{Evaluation Metrics: Rouge Scores}
ROUGE  is a standard evaluation measure for summarization tasks, and it stands for Recall-Oriented Understudy for Gisting Evaluation. This metric was initially introduced in the paper \cite{Chin-2004} and considers both Precision and Recall between predicted and reference summaries. It includes three scores: ROUGE-1, ROUGE-2, and ROUGE-L. 
\begin{itemize}
    \item \textbf{Recall (R):} Recall calculates the total number of n-grams present in the predicted summary and the
reference and then it gets divided by the total number of n-grams present in the
reference summary.
\[R=\frac{Count_{match} (gram_n) }{Count(ref-gram_n)}\]
\item \textbf{Precision (P):} It is calculated in almost the exact same way like Recall, but rather than dividing by the reference n-gram count, we divide by the predicted n-gram count.
 \[P=\frac{Count_{match} (gram_n) }{Count(cand-gram_n)}\]
 \item \textbf{F1-Score(F):}  That gives us a reliable measure of our model performance that relies not only on the model capturing as many words as possible (recall) but doing so without outputting irrelevant words (precision).
 \[F1-Score(F)=2*\frac{precision\ *\ recall}{precision\ +\ recall}\]
 \item \textbf{Rouge1, Rouge2, Rouge L Scores:}
Rouge-1 matches the number of Uni grams with our machine generated summary and with
the referenced summary. And in the case of Rouge-2 it matches the bi-grams. Rouge-L is used to measure the longest sequences of sharing tokens between our reference summary and the machine generated summary.
\end{itemize}

\subsection{BERT Score}
BERT Score \cite{Tiyani2019} is an evaluation metric for text generation tasks. We evaluate our candid-reference summaries by computing BERT Score. For each token $x$ in reference summary, the BERT embedding is denoted as $\textbf{x}.$ Similarly, for each token $\hat{x}$ in candid summary, the BERT embedding is denoted as $\hat{\textbf{x}}.$ For given sentence, assume that that {$\textbf{x}=\langle \textbf{x}_1,\ldots,\textbf{x}_m\rangle$} and \textbf{$\hat{\textbf{x}}=\langle \hat{\textbf{x}}_1,\ldots,\hat{\textbf{x}}_\ell\rangle$} are BERT embedding of reference $x$ and candid $\hat{x}$ summaries respectively. Recall, precision, and F1 scores for $x$ and $\hat{x}$ are defined as follows:
\[R_{BERT}={1\over|x|}\sum_{x_i\in x}max_{\hat{x}_j\in \hat{x}}\textbf{x}_i^T\hat{\textbf{x}}_j,\ P_{BERT}={1\over|\hat{x}|}\sum_{\hat{x}_j\in \hat{x}}max_{x_i\in x}\textbf{x}_i^T\hat{\textbf{x}}_j,\]
\[F_{BERT}=2{P_{BERT}.R_{BERT}\over P_{BERT}+R_{BERT}}\]

\subsection{Results}
The following values are observed in our experiments for smaller documents:
\begin{itemize}
    \item \textbf{Rouge-1 and Rouge-2 Scores:}\\
    \[
    \begin{tabular}{|c| c| c |c|| c| c |c| }
\hline
 &\multicolumn{3}{|c|}{\textbf{Rouge-1 Score}}&\multicolumn{3}{|c|}{\textbf{Rouge-2 Score}} \\
 \hline
Model & recall & precision & f1-score&recall & precision & f1-score\\
 \hline
 Tf-Idf & 0.49 & 0.64 & 0.55&0.46&0.57&0.51\\ 
 %Fast-Text & 0.33 & 0.33 & 0.33 \\  
 BERT (Vanila) & 0.51 & 0.58 & 0.54&0.46&0.52&0.49 \\
 San-BERT & \textbf{0.61} & 0.58 & 0.59&0.54&0.52&0.53 \\ 
 San-ALBERT & 0.38 & \textbf{0.71} & \textbf{0.69}&\textbf{0.66}6&\textbf{0.64}&\textbf{0.65}\\ 
 San-RoBERTa & {0.52} & 0.52 & {0.51}&{0.46}&0.46&0.46 \\ 
 \hline
\end{tabular}
\]
    \item \textbf{Rouge-L and BERT Scores:}
        
\begin{center}
\begin{tabular}{|c| c| c |c|| c| c |c|}
\hline
 Model&\multicolumn{3}{|c||}{\textbf{Rouge-L Score}}&\multicolumn{3}{|c|}{\textbf{BERT Score}} \\
 \hline
& recall & precision & f1-score& recall & precision & f1-score\\
 \hline
 Tf-Idf & 0.48 & 0.62 & 0.54& 0.892 & 0.921 & 0.906\\ 
 %Fast-Text & 0.13 & 0.13 & 0.13 \\  
 BERT (Vanila) & 0.48 & 0.56 & 0.52& 0.904 & 0.913 & 0.908 \\
 San-BERT & 0.60 & 0.56 & 0.57& 0.921 & 0.914 & 0.917 \\ 
 San-ALBERT & \textbf{0.69} & \textbf{0.70} & \textbf{0.68} & \textbf{0.937} & \textbf{0.942} & \textbf{0.939} \\ 
 San-RoBERTa & 0.50 & 0.51 & 0.49& 0.899 & 0.907 & 0.902 \\ 
 \hline
\end{tabular}    
\end{center}
\end{itemize}

The following values are observed in our experiments for longer documents:\\
\begin{itemize}
    \item \textbf{Rouge-1 and Rouge-2 Scores:}\\
\[
\begin{tabular}{|c| c| c |c|| c| c |c| }
\hline
 &\multicolumn{3}{|c|}{\textbf{Rouge-1 Score}}&\multicolumn{3}{|c|}{\textbf{Rouge-2 Score}} \\
 \hline
Model & recall & precision & f1-score& recall & precision & f1-score\\
 \hline
 Tf-Idf & 0.46 & 0.59 & 0.52& 0.40 & 0.47 & 0.43\\ 
 %Fast-Text & 0.33 & 0.33 & 0.33 \\  
 Bert (Vanila) & 0.59 & 0.65 & 0.62& 0.54 & 0.56 & 0.55 \\
 San-BERT & 0.72 & 0.72 & 0.71& 0.66 & 0.64 & 0.64 \\ 
 San-ALBERT & \textbf{0.75} & \textbf{0.77} & \textbf{0.76}& \textbf{0.70} & \textbf{0.70} & \textbf{0.70} \\ 
 San-RoBERTa & \textbf{0.50} & 0.58 & \textbf{0.54}& {0.46} & 0.49 &{0.47} \\ 
 \hline
\end{tabular}
\]
    \item \textbf{Rouge-L and BERT Scores:}\\
\end{itemize}
\begin{center}
\begin{tabular}{|c| c| c |c|| c| c |c| }
\hline
 &\multicolumn{3}{|c|}{\textbf{Rouge-L Score}}&\multicolumn{3}{|c|}{\textbf{BERT Score}} \\
 \hline
Model & recall & precision & f1-score& recall & precision & f1-score\\
 \hline
 Tf-Idf & 0.41 & 0.52 & 0.46& 0.893 & 0.924 & 0.907\\ 
 %Fast-Text & 0.13 & 0.13 & 0.13 \\  
 Bert (Vanila) & 0.57 & 0.63 & 0.60& 0.932 & 0.933 & 0.933 \\
 San-BERT & \textbf{0.704} & 0.707 & 0.70& 0.941 & 0.929 & 0.935 \\ 
 San-ALBERT & 0.701 & \textbf{0.713} & \textbf{0.707}&  \textbf{0.949} & \textbf{0.962} &  \textbf{0.955} \\ 
 San-RoBERTa & {0.50} & 0.58 & {0.54}& {0.909} & 0.927 & {0.918} \\ 
 \hline
\end{tabular}    
\end{center}

\section{Conclusions and Future Works}
Our extractive text summarization models performed admirably. The results obtained using the San-ALBERT and San-BERT models are superior to those obtained using the San-RoBERTa model and the TF-IDF method. These results may differ depending on the documents and their related summaries; the referenced summary we utilized for evaluation was human-generated. Compared to other extraction processes used in the past, the results are generally favorable. In future work \cite{Kunal22}, the authors develop an abstract summarization model for Sanskrit documents using the San-BERT and Point Generator models.
\section{Acknowledgements}
We dedicate this work to Bhagawan Sri Sathya Sai Baba.



\bibliographystyle{unsrt}  
\bibliography{bibliography}  %%% Remove comment to use the external .bib file (using bibtex).

%\begin{thebibliography}{1}

\end{document}